\documentclass[11pt]{article}

\usepackage[final]{acl}

\usepackage{times}
\usepackage{latexsym}

\usepackage[T1]{fontenc}

\usepackage[utf8]{inputenc}
\usepackage{pgfplots}
\pgfplotsset{compat=1.18}
\usepackage{microtype}

\usepackage{inconsolata}
\usepackage{listings}
\usepackage{amssymb}

\usepackage{graphicx}
\usepackage{tikz}
\usetikzlibrary{shapes.geometric, shapes.misc, arrows.meta, positioning, fit, backgrounds, calc, decorations.pathreplacing}
\usepackage{booktabs}   
\usepackage{amsmath}    
\usepackage{tcolorbox} 
\usepackage[table]{xcolor}
\usepackage{acronym} 

\definecolor{lookup}{HTML}{E8F4FD}
\definecolor{schema}{HTML}{FFF3E0}
\definecolor{gen}{HTML}{E8F5E9}
\definecolor{exec}{HTML}{FCE4EC}
\definecolor{reflect}{HTML}{F3E5F5}

\acrodef{FLINT}[FLINT]{Financial LINking Text-to-SQL}
\acrodef{T2S}[Text-to-SQL]{Text-to-SQL}
  
\lstdefinestyle{sql}{
    language=SQL,
    basicstyle=\ttfamily\small,
    keywordstyle=\bfseries,
    breaklines=true
  }

\newcommand{\cmark}{\checkmark}
\newcommand{\xmark}{\texttimes}

\title{From Benchmarks to Production: A Text-to-SQL System for Complex Financial Data}

\author{
 \textbf{Arijit Sehanobish\thanks{Equal Contribution}},
 \textbf{Bruno Gomes Coelho\footnotemark[1]},
 \textbf{Guillaume Michel},
 \textbf{ Sophia Zhi},
\\
 \textbf{Valerie Faucon-Morin},
 \textbf{Kristen Howell}
\\
Kensho Technologies
\\
 \small{
   \textbf{Correspondence:} \href{mailto:email@domain}{arijit.sehanobish@kensho.com}
 }
}

\begin{document}
\maketitle
\begin{abstract}
General-purpose \acs{T2S} systems achieve strong performance on academic benchmarks like Spider and BIRD, where schemas are relatively shallow and column values are often human readable. In production financial databases, where concepts are stored as opaque integer keys rather than human-readable strings, these methods fall below 50\%, as even simple queries require multiple joins and filter predicates reference opaque IDs. We present \ac{FLINT}, a domain-specialized \acs{T2S} system that closes this gap through
  three key components: (1) a lookup agent that dynamically resolves natural-language concepts to question-specific reference table constraints, (2)
  embedding-based retrieval of structurally similar query templates from a compact, expert-authored bank, and (3) schema linking that prunes a large table schema to the relevant subset by traversing foreign-key chains, rather than relying on name similarity alone. We evaluate on two datasets totaling 359 questions over production financial schemas. \ac{FLINT} outperforms various state-of-the-art baselines using the same LLM. The system is deployed in production as part of a financial data retrieval service.\footnote{\href{https://kensho.com/solutions/adaptive-retrieval}{https://kensho.com/solutions/adaptive-retrieval}}

\end{abstract}

\section{Introduction}

\acs{T2S}, translating natural language questions into executable database queries, has emerged as a useful interface for non-technical users to access structured data. Recent systems have achieved 70-85\% accuracy on academic benchmarks such as Spider~\cite{yu-etal-2018-spider} and BIRD~\cite{li2024bird} through advances in schema linking, multi-agent architectures, and prompt engineering.

However, these results do not transfer to real-world database schemas in the financial sector. Questions about these databases 
(e.g., a company's quarterly revenue, the largest M\&A deals in a sector) 
are straightforward to express in natural language, but difficult to translate into production-ready SQL.
Commercial financial databases optimize for storage efficiency, not human readability, so financial metrics, period types, and currencies are encoded as integer foreign keys across dozens of normalized reference tables. A single time-series question requires traversing a six-table join chain, and the filters that distinguish ``annual revenue'' from ``quarterly net income'' are opaque integers that cannot be inferred from column names alone (detailed examples in App.~\ref{subsec:appendix:financials_dataset} and~\ref{subsec:appendix:transactions_dataset}).

This setting exposes a gap that academic benchmarks do not measure. In Spider~\cite{yu-etal-2018-spider} and BIRD~\cite{li2024bird}, the column values are self-describing (e.g. \textit{country = `USA'}), the joins are shallow (1-2 on average), and the schemas fit into a single prompt. 
We evaluate seven state-of-the-art methods on two production financial datasets and find that none exceeds 50\% accuracy, despite generating SQL with over 99\% execution success. While the queries run, they often return the wrong data due to incorrect reference table filter values. 

We present \acf{FLINT}, a domain-specialized \acs{T2S} system that targets this through three components: 
(1) a lookup agent that dynamically resolves natural-language concepts to question-specific reference table constraints, (2) embedding-based retrieval over a compact, expert-authored query bank for structural templates, and (3) schema linking that prunes large schemas to relevant subsets, including the bridge tables needed for foreign-key traversal.

This design decouples domain knowledge from model weights.
When schemas evolve, new metrics are added, or coverage gaps are identified, domain experts update the configuration artifacts directly without retraining or ML engineering involvement.
Our architecture addresses normalized schemas that are common across industries, and while we focus on financial data, we validate the generalizability of this approach in additional scientific and industrial domains in App.~\ref{sec:generalization}.

To summarize, our contributions are:
\begin{itemize}

\item \textbf{Bottleneck diagnosis.} On normalized production-style schemas, value grounding and not SQL generation is the dominant failure mode: 76--96\% of errors are wrong reference-table filter IDs.

\item \textbf{A lightweight value-grounding method.} We introduce a model-independent configuration layer that targets this bottleneck. On four public benchmarks, it outperforms CHESS and ReFoRCE on three datasets at $5\times$ lower generation cost.

\item \textbf{Semantic evaluation.} We introduce a three-tier evaluation pipeline (AST canonicalization $\to$ column-aligned comparison $\to$ LLM judge) that distinguishes semantically equivalent queries from genuine errors.

\end{itemize}

\section{Related Work}



Prompt engineering approaches to \acs{T2S}~\cite[][further discussion in App.~\ref{sec:appendix:expanded_related_work}]{gao2024dail,pourreza2024din} have largely been superseded by multi-agent decomposition, where SQL generation is split across multiple specialized stages before the final query is produced. Recent systems address different failure modes: candidate diversity and selection~\cite{talaei2024chess,pourreza2025chase}, interactive database exploration~\cite{wang2024macsql,flexsql}, and iterative refinement via execution feedback~\cite{reforce}. While these strategies reduce structural SQL errors, none includes a dedicated mechanism for resolving opaque reference-table IDs. On normalized schemas in finance, this shared gap is the main source of error (see Section~\ref{sec:baselines}).

Evaluation methodology and benchmarks remain active research concerns. 
Spider~\cite{yu-etal-2018-spider} compares SQL components structurally via exact set match, rejecting equivalent reformulations.
In contrast, BIRD~\cite{li2024bird} looks only at query outputs, allowing formatting differences to be conflated with semantic errors. This is a limitation that SpotIt~\cite{klopfenstein2026spotit} quantifies through formal verification, hypothesizing that up to 20\% of BIRD evaluation data may be incorrect because test databases coincidentally agree on non-equivalent queries.
Our benchmarks differ from Spider and BIRD in both content and evaluation: The schemas feature deep normalization with opaque IDs, mandatory multi-table joins for even simple queries, and implicit domain conventions absent from the natural language.  Furthermore, we evaluate through a three-tier pipeline (AST canonicalization, output comparison, and LLM judge) designed to accommodate the many valid reformulations that arise on such schemas. We discuss additional related work in App.~\ref{sec:appendix:expanded_related_work}.

\section{System Architecture}
\label{sec:architecture}

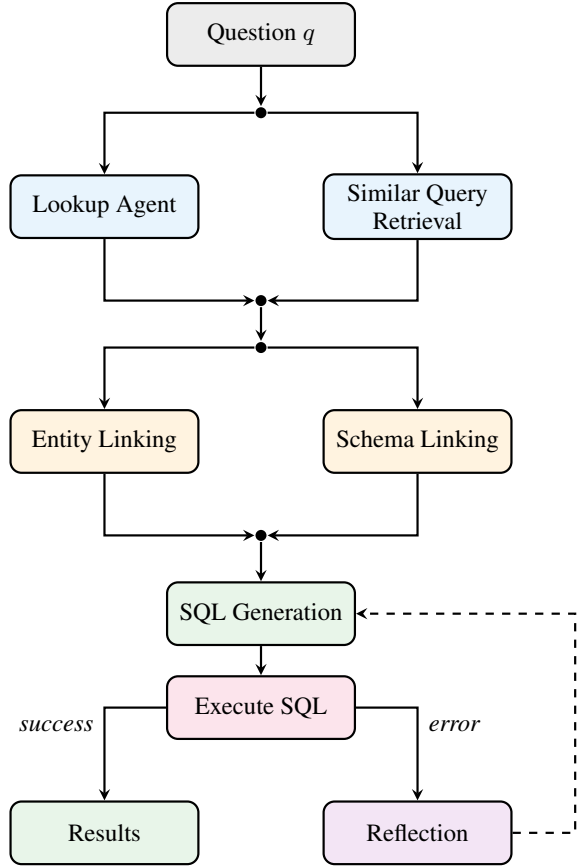
\begin{figure}[t]
\centering
\resizebox{\columnwidth}{!}{%
\begin{tikzpicture}[
    node distance=1.0cm and 2.0cm,
    every node/.style={font=\small},
    block/.style={rectangle, draw, rounded corners, minimum width=2.4cm, minimum height=0.8cm, align=center},
    dot/.style={circle, fill=black, inner sep=0pt, minimum size=4pt},
    >=stealth,
    thick
]

\node[block, fill=gray!15] (input) at (0,0) {Question $q$};

\node[dot] (fork1) at (0,-1.0) {};
\node[block, fill=lookup] (lookupnode) at (-2.0,-2.2) {Lookup Agent};
\node[block, fill=lookup] (retrieval) at (2.0,-2.2) {Similar Query\\Retrieval};

\node[dot] (join1) at (0,-3.4) {};

\node[dot] (fork2) at (0,-4.0) {};
\node[block, fill=schema] (entity) at (-2.0,-5.2) {Entity Linking};
\node[block, fill=schema] (schemalinking) at (2.0,-5.2) {Schema Linking};

\node[dot] (join2) at (0,-6.4) {};

\node[block, fill=gen] (sqlgen) at (0,-7.4) {SQL Generation};
\node[block, fill=exec] (execute) at (0,-8.6) {Execute SQL};

\node[block, fill=gen] (results) at (-2.0,-10.2) {Results};
\node[block, fill=reflect] (reflection) at (2.0,-10.2) {Reflection};

\draw[->] (input) -- (fork1);

\draw[->] (fork1) -| (lookupnode);
\draw[->] (fork1) -| (retrieval);

\draw[->] (lookupnode) |- (join1);
\draw[->] (retrieval) |- (join1);

\draw[->] (join1) -- (fork2);

\draw[->] (fork2) -| (entity);
\draw[->] (fork2) -| (schemalinking);

\draw[->] (entity) |- (join2);
\draw[->] (schemalinking) |- (join2);

\draw[->] (join2) -- (sqlgen);
\draw[->] (sqlgen) -- (execute);

\draw[->] (execute) -| node[pos=0.6, left, font=\footnotesize] {\textit{success}} (results);
\draw[->] (execute) -| node[pos=0.6, right, font=\footnotesize] {\textit{error}} (reflection);

\draw[->, dashed] (reflection.east) -- ++(0.8,0) |- (sqlgen.east);

\end{tikzpicture}
}
\caption{\ac{FLINT} pipeline. Stages arranged horizontally execute in parallel. Dashed arrows show the error-driven reflection loop.}
\label{fig:architecture}
\end{figure}

\ac{FLINT} decomposes natural language to SQL generation into a pipeline of specialized stages, each addressing a distinct sub-problem.
Given a user question $q$, the system produces a SQL query $\hat{s}$ through six stages: lookup, similar query retrieval, entity linking, schema linking, SQL generation, and execution with reflection.
Stages without data dependency -- lookup and similar query retrieval, followed by entity linking and schema linking -- execute in parallel (Fig.~\ref{fig:architecture}). This parallelism yields median end-to-end latency of 10 - 20s (App.~\ref{sec:latency_analysis}).

\subsection{Lookup Agent}
\label{sec:lookup}

Financial databases encode categorical information like industries, period types, and financial metrics as numeric IDs in reference tables.
A question mentioning ``quarterly EBITDA'' requires knowing which integer ID maps to ``Quarterly'' and which maps to ``EBITDA''. This cannot be inferred from the schema, which shows only that these are integer foreign keys (Fig~\ref{fig:opaque_ids}). The Lookup Agent resolves these references in two sub-steps:

  \begin{enumerate}
      \item \textbf{Table selection:} An LLM determines which reference tables are relevant to the question. A question about ``Canadian companies'' triggers a 
  geography table lookup; ``revenue'' triggers the financial metrics table.
      \item \textbf{Constraint generation:} The agent executes \texttt{SELECT} queries against the identified production reference table(s), presents retrieved rows to the LLM, and asks it to produce the appropriate SQL filters. On Financials, metric resolution is a two-stage funnel: the full 11{,}412-item metrics table is first reduced to an SME-curated catalog of ${\sim}2{,}500$ commonly-used items (the remaining ${\sim}9{,}000$ are rarely queried, and the curation is itself domain knowledge), and a per-question fuzzy-matching step then narrows this to the ${\sim}50$--$100$ candidates from which the LLM selects. The Transactions reference tables are small and passed whole.
  \end{enumerate}
  
On our evaluation set, this pre-filtering maintains 98\% recall (measured against gold query).
The output is a set of \textbf{lookup constraints}: SQL \texttt{WHERE}-clause fragments passed to subsequent stages.

\subsection{Similar Query Retrieval}
\label{sec:retrieval}

\ac{FLINT} maintains a \textbf{query bank} of expert-authored examples, each containing a question, reasoning traces, and reference SQL. At inference, the $k{=}3$ most similar examples are retrieved via cosine similarity over a vector store.

Each example is hand-crafted by a domain expert to cover a diverse set of distinct query patterns (e.g., single-entity lookup,\footnote{Query bank examples use a canonical entity (a fixed reference company) to ensure similarity captures the \textit{type} of data retrieval rather than surface-level entity overlap.} multi-period comparison, cross-currency ranking). 
The query bank and evaluation set are strictly disjoint. Retrieved examples serve two roles: \textit{structural templates} for schema linking (showing correct join patterns) and \textit{SQL exemplars} for generation (demonstrating syntax and filter placement).

\subsection{Entity Linking}
\label{sec:entity}

Entity linking resolves natural-language mentions (company names, ticker symbols) to the opaque integer keys used in the database.
Without this step, the model has no path from a mention like ``Nvidia'' to the correct numeric identifier, as the core data tables contain no human-readable name column, and fuzzy string matching against 20 million+ entities is unreliable at production scale. \ac{FLINT} delegates resolution to an external entity disambiguation service, Kensho Link,\footnote{\href{https://kensho.com/solutions/core-ai-capabilities\#link}{https://kensho.com/solutions/core-ai-capabilities\#link}} that handles name variations and identifier-type disambiguation, returning canonical integer IDs. 
Resolved IDs are injected as concrete \texttt{WHERE}-clause constraints into the generation prompt.

\subsection{Schema Linking}
\label{sec:schema_linking}

Schema linking determines which tables, columns, join conditions, and filter predicates are required for the query. Given the question, reference table constraints, lookup IDs, and retrieved examples, the LLM selects the relevant subset of the schema. This is challenging because the answer rarely lives in a single table and reaching it requires identifying intermediate tables that serve only as join bridges. For example, retrieving a
  stock price for a company requires resolving the company's security, then its trading item, before reaching the price table: \texttt{Company} $\rightarrow$
  \texttt{Security} $\rightarrow$ \texttt{TradingItem} $\rightarrow$ \texttt{PriceEquity}.

The prompt includes:
(1) the full database schema with column descriptions and documented join paths, 
(2) lookup constraints from the Lookup Agent (resolving which integer IDs to filter on), 
(3) domain-specific instructions encoding implicit conventions (e.g., default temporal granularity, which of several price columns to use, required flag filters for primary listings), and
(4) retrieved examples with their reasoning traces.

The LLM outputs a structured representation consisting of selected columns, join conditions, and filters. This decouples \textit{what to query} from \textit{how to write SQL}: Schema linking reasons about table selection while generation handles syntax. Since schema linking sees the full schema, this design suits our target of a fixed,
curated schema.

\subsection{SQL Generation}
\label{sec:generation}

Given schema links, entity IDs, and SQL exemplars, the LLM generates the final query.
Domain constraints are enforced via the prompt: Use only specified tables and columns, apply entity IDs as concrete filters, include contextual columns (company name, currency) for interpretability, and follow the target SQL dialect. Before execution, a safety validator ensures the query only accesses allowed schemas and tables, preventing unintended data exposure.

\subsection{Execution and Reflection}
\label{sec:execution}

The query is executed against the database.
On execution success, results are returned directly.
On failure, the system enters a \textbf{reflection loop}: The LLM receives the original question, failed SQL, and database error, then regenerates with targeted corrections. This reflection step is particularly useful for complex schemas, where execution errors often stem from localized mistakes, such as missing join conditions, incorrect aliases, or column typos, which become diagnosable once the error message is available. While we allow only one reflection attempt to stay within production latency constraints, this is configurable for use cases where accuracy is prioritized over response time.

\subsection{Dataset Configuration}
\label{sec:configuration}

\ac{FLINT} adapts to new datasets without fine-tuning or retraining.
Configuration requires four artifacts:
\begin{enumerate}
    \item A schema CSV mapping tables and columns to descriptions and join-path documentation.
    \item A list of reference tables for the Lookup
Agent (identifying tables with categorical IDs).
    \item A query bank of 15--30 examples covering major query patterns. 
    \item Domain-specific instruction prompts for schema linking and generation (encoding implicit conventions).
\end{enumerate}
Because these artifacts are structured domain knowledge, they are authored by subject-matter experts rather than ML engineers. In our experience a moderate-complexity dataset takes roughly one person-week to configure, and a complex one (many reference tables, several implicit conventions) takes two to three weeks including iterative refinement; adding a new example for a novel pattern
is ${\sim}30$ minutes (App.~\ref{sec:appendix_config}). The pipeline is general: Adapting \ac{FLINT} to a new financial dataset requires only new configuration artifacts, not implementation changes. FLINT's deployment details, including caching, fallback strategies, and latency constraints, is presented in App.~\ref{sec:appendix_deployment}.

\section{Experiments}
\label{sec:experiments}
We first describe our proprietary datasets (Sec.~\ref{sec:dataset}), our multi-stage evaluation protocol (Sec.~\ref{sec:evaluation}), compare our results with various well known multi-agent systems~(Sec.~\ref{sec:baselines}), and detailed ablation studies (Sec.~\ref{sec:ablation}), and error analysis~(Sec.~\ref{sec:error_analysis}). 
\subsection{Datasets}
\label{sec:dataset}

We evaluate \ac{FLINT} on two datasets constructed over a large-scale financial database with deep schema normalization, summarized in Table~\ref{tab:schema_stats}.

The Financials dataset covers company intelligence, market data, financial metrics, and securities across global public companies.
The schema consists of 24 tables (8 core data tables, 16 reference tables). Financial values are not stored with human-readable labels but metric names, period types, unit types, and currencies are each encoded as integer foreign keys into separate reference tables.
A typical query requires joining 6 tables; complex screening queries require up to 13 joins.
The dataset contains 221 natural-language questions authored by domain experts, covering company fundamentals (44\%), market data (14\%), market capitalization (10\%), screening/ranking (8\%), industry classification (7\%), and other mixed queries (17\%).

The Transactions dataset covers M\&A activity, funding rounds, stock buybacks, and corporate relationships.
The schema consists of 18 tables (6 core, 12 reference/lookup) capturing deal types, advisor roles, company relationships, and transaction statuses.
Queries require joining 5 tables on average to resolve entity relationships and transaction details.
The dataset contains 138 natural-language questions authored by domain experts, covering acquisitions (50\%), stock buybacks (14\%), general transaction screening (14\%), bankruptcies (8\%), spin-offs and divestitures (6\%), investments and funding (4\%), and other deal types (4\%).
\begin{table}[h]
\centering
\caption{\small{Schema and benchmark statistics for both datasets. Reference tables store lookup values (metrics, period types, currencies); core tables store financial data and transactions.}}
\small
\setlength{\tabcolsep}{6pt}
\renewcommand{\arraystretch}{1.15}
\begin{tabular}{@{}l r r@{}}
\toprule
\textbf{Property} & \textbf{Financials} & \textbf{Transactions} \\
\midrule
Total tables & 24 & 18 \\
\quad Reference / lookup & 16 & 12 \\
\quad Core data & 8 & 6 \\[3pt]
Avg.\ joins per query & 6.0 & 5.2 \\
Max joins (single query) & 13 & 13 \\
Queries requiring CTEs & 40 & 17 \\[3pt]
Evaluation questions & 221 & 138 \\
\bottomrule
\end{tabular}
\label{tab:schema_stats}
\end{table}

The 359 evaluation questions are drawn from a pool of production queries logged over 6 months, deduplicated and filtered to ensure coverage across query categories, table usage, and difficulty levels. Each question is paired with a gold SQL query written by database domain experts.\footnote{This may bias the benchmark toward questions with known SQL solutions, which we consider acceptable for measuring system coverage of production use cases.} The gold queries enforce constraints central to correctness on normalized schemas: explicit joins, required flag filters for selecting primary listings, currency normalization for cross-country comparisons, and correct temporal-granularity constraints.
These are precisely the constraints that general-purpose \acs{T2S} systems fail to satisfy (Sec.~\ref{sec:baselines}). While we focus on the financial domain, the constraints we address are present in other domains. We validate the generalizability of our system on two external benchmarks spanning various scientific and industrial domains in App.~\ref{sec:generalization}.

The query bank used for few-shot retrieval (27 for Financials, 18 for Transactions) (Section~\ref{sec:retrieval}) is disjoint from the evaluation set: No evaluation question appears in, was derived from, or shares an entity with any query bank example. To quantify residual similarity, we measure the normalized tree-edit distance
(TED) from each evaluation query to its nearest query-bank neighbor over \texttt{sqlglot} ASTs (overlap $=1-\text{TED}$). Content-aware overlap is small with no near-duplicate on Transactions, so the query bank supplies structural patterns and not answers (App.~\ref{sec:appendix_overlap}).

\subsection{Evaluation}
\label{sec:evaluation}

SQL queries that are syntactically different can produce identical results, while structurally similar queries can return different data due to subtle filter differences.
We use a three-tier evaluation pipeline:

\begin{enumerate}
    \item \textbf{Execution:} We execute both gold and generated queries against the production database; execution failures score 0 immediately.
    \item \textbf{Heuristics:} We parse both queries into abstract syntax trees using \texttt{sqlglot}, apply canonicalizing transformations (alias normalization, date function standardization, JOIN reordering), and compute tree edit distance. Queries that differ only in aliasing, column order, or semantically equivalent reformulations (e.g., CTE vs.\ subquery) are matched. For queries that are not matched via AST, gold columns are checked against the generated columns via case-insensitive name alignment, and values are compared with order-invariant set equality. This heuristic resolves almost 60\% of cases at negligible cost.
    \item \textbf{LLM Judge:} We prompt an LLM to determine semantic equivalence by comparing result tables with reference to the original question. The LLM judge decides only ${\sim} 40 \%$ of comparisons (the rest are settled by heuristics or execution), and on a validated sample it agrees with human judgments ${\geq}94\%$ of the time, bounding a $\pm3$-point worst case on the reported accuracies. 
\end{enumerate}

\noindent We report accuracy as the percentage of questions whose gold and generated SQL are marked equivalent in step 2 or 3 (additional details in App.~\ref{sec:appendix_ast}). 

\begin{figure}
    \centering
    \resizebox{\columnwidth}{!}{%
    \begin{tikzpicture}[
    every node/.style={font=\small},
    mainblock/.style={rectangle, draw=blue!70, line width=1.2pt, rounded corners=2pt, minimum width=3.2cm, minimum height=1.6cm, align=center, fill=blue!5},
    prepblock/.style={rectangle, draw=gray!50, rounded corners=4pt, minimum width=2.8cm, minimum height=1.0cm, align=center, fill=gray!5},
    scoreblock/.style={rectangle, draw=gray!50, rounded corners=4pt, minimum width=2.2cm, minimum height=1.0cm, align=center, fill=gray!5},
    >=stealth,
    thick
]

\node[prepblock] (parse) at (0, 2.0) {Parse gold \&\\generated SQL};

\node[mainblock] (ast) at (0, 0) {\textbf{AST Comparison}\\[3pt]{\footnotesize\textit{sqlglot, tree edit distance}}};
\node[mainblock] (output) at (5.0, 0) {\textbf{SQL Output Comparison}\\[3pt]{\footnotesize\textit{column alignment, set equality}}};
\node[mainblock] (llm) at (10.0, 0) {\textbf{LLM Judge}\\[3pt]{\footnotesize\textit{semantic equivalence}}};

\node[scoreblock] (score) at (5.0, -3.0) {\textbf{Score}};

\draw[->] (parse.south) -- (ast.north);

\draw[->] (ast.east) -- node[above, font=\footnotesize] {mismatch} (output.west);
\draw[->] (output.east) -- node[above, font=\footnotesize] {mismatch} (llm.west);

\draw[->, dashed, green!60!black] (ast.south) -- node[left, font=\footnotesize, green!60!black, pos=0.4] {match} (score.west);
\draw[->, dashed, green!60!black] (output.south) -- node[right, font=\footnotesize, green!60!black, pos=0.5] {match} (score.north);
\draw[->, dashed, green!60!black] (llm.south) -- node[right, font=\footnotesize, green!60!black, pos=0.4] {verdict} (score.east);

\end{tikzpicture}
    }
    \caption{\small{Our multi-stage evaluation pipeline: The generated queries that execute successfully are first matched via AST trees with the gold query. For the unmatched queries we pass them through heuristics and eventually to LLM-as-a-judge.}}
    \label{fig:eval_pipeline}
\end{figure}
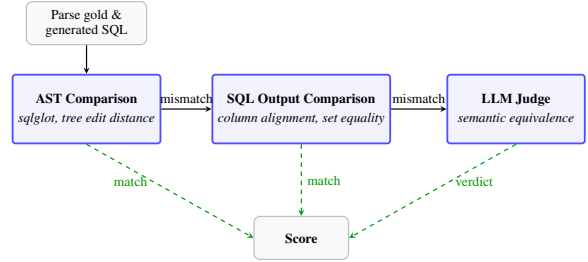

\subsection{Comparison with Existing Methods}
\label{sec:baselines}

We compare against seven \acs{T2S} systems spanning multi-agent architectures (CHESS~\cite{talaei2024chess}, Chase-SQL~\cite{pourreza2025chase}, MAC-SQL~\cite{wang2024macsql}, FlexSQL~\cite{flexsql}, ReFoRCE~\cite{reforce}) and prompt-engineering approaches (DAIL-SQL~\cite{gao2024dail}, DIN-SQL~\cite{pourreza2024din}) (Tab.~\ref{tab:baselines}).
All use Claude 4.5 Opus~\citep{anthropic2025opus}.

\begin{table*}[t]
\centering
\caption{\small{Comparison with existing \acs{T2S} methods on normalized financial schemas.
All methods use Claude 4.5 Opus and share base infrastructure (App.~\ref{subsec:appendix_baselines-shared_infra}); DAIL-SQL, DIN-SQL, and FlexSQL additionally receive shared pipeline components (App.~\ref{subsec:appendix_baselines-summary_adaptations}), making their comparison conservative; execution success is $\geq$99\% for all.
}} 
\small
\setlength{\tabcolsep}{6pt}
\renewcommand{\arraystretch}{1.2}
\begin{tabular}{@{}l l r r@{}}
\toprule
& & \textbf{Fin.} & \textbf{Trans.} \\
\textbf{Method} & \textbf{Approach} & \multicolumn{2}{c}{\footnotesize\textit{Accuracy\ (\%)}} \\
\midrule
\rowcolor{blue!6}
\textbf{\ac{FLINT}} & \textit{Domain-specialized agents} & \textbf{68.8} & \textbf{65.2} \\
\addlinespace[6pt]
\multicolumn{4}{@{}l}{\footnotesize\textsc{Multi-agent / multi-candidate}} \\
\addlinespace[2pt]
CHESS~\cite{talaei2024chess} & Self-consistency ($k{=}1$) & 27.1 & 42.0 \\
CHESS~\cite{talaei2024chess} & Self-consistency ($k{=}10$) & 35.6 & 49.6 \\
Chase-SQL~\cite{pourreza2025chase} & Multi-path + tournament & 16.3 & 21.0 \\
MAC-SQL~\cite{wang2024macsql} & Agentic decomposition & 33.3 & 37.3 \\
FlexSQL~\cite{flexsql} & Plan + schema explore & 43.4 & 40.6 \\
ReFoRCE~\cite{reforce} & Iterative refinement & 46.6 & 47.1 \\
\addlinespace[6pt]
\multicolumn{4}{@{}l}{\footnotesize\textsc{Prompt engineering}} \\
\addlinespace[2pt]
DAIL-SQL~\cite{gao2024dail} & Few-shot selection & 31.7 & 39.9 \\
DIN-SQL~\cite{pourreza2024din} & Decompose + classify & 33.5 & 36.5 \\
\addlinespace[6pt]
\multicolumn{4}{@{}l}{\footnotesize\textsc{Direct prompting (entity linking only)}} \\
\addlinespace[2pt]
Claude 4.5 Opus & Zero-shot & 12.7 & 15.9 \\
Claude 4.5 Opus & Few-shot ($k{=}3$) & 38.0 & 42.8 \\
\bottomrule
\end{tabular}
\label{tab:baselines}
\end{table*}

All methods produce valid SQL ($\geq$99\% execution success) but fail semantically: The dominant error is syntactically correct queries with wrong filter values. No baseline exceeds 50\% accuracy, despite achieving substantially higher performance on Spider/BIRD where column values are often semantically informative (e.g.~\texttt{country = `USA'}).
On our schemas, mapping from natural language (``annual revenue'') to opaque integer IDs requires querying reference tables that no baseline provides.
CHESS $k{=}10$ gains 8.5 points over $k{=}1$ on Financials via sampling diversity (occasionally hitting correct IDs), but self-consistency cannot fix errors shared across all candidates.
Among baseline failures, 64--96\% contain incorrect IDs (App.~\ref{sec:appendix_baselines}). These gaps are well outside sampling noise: Wilson 95\% intervals on FLINT are
$68.8\%\,[62.4,74.5]$ (Financials) and $65.2\%\,[56.9,72.6]$ (Transactions), and the margin over the strongest baseline exceeds them by a wide margin; the judge's $\pm4$-point worst case cannot close a gap this size. Moreover, FLINT is the fastest system evaluated, completing queries in 10--20s median latency, faster than all seven baselines (Fig.~\ref{fig:latency}).

\paragraph{Generalization to other domains.} Moreover, to test domain generality, we evaluate FLINT on two external benchmarks spanning medical, scientific, and public-sector domains; with only minor configuration changes (schema documentation, query bank, and domain rules), FLINT outperforms both CHESS and ReFoRCE while remaining almost 5$\times$ faster. App.~\ref{sec:generalization} provides further discussion, additional results on SM3~\cite{sivasubramaniam2024smtexttoquery} and comparisons with few-shot prompting.
\begin{figure}[h]
\centering
\includegraphics[width=\linewidth]{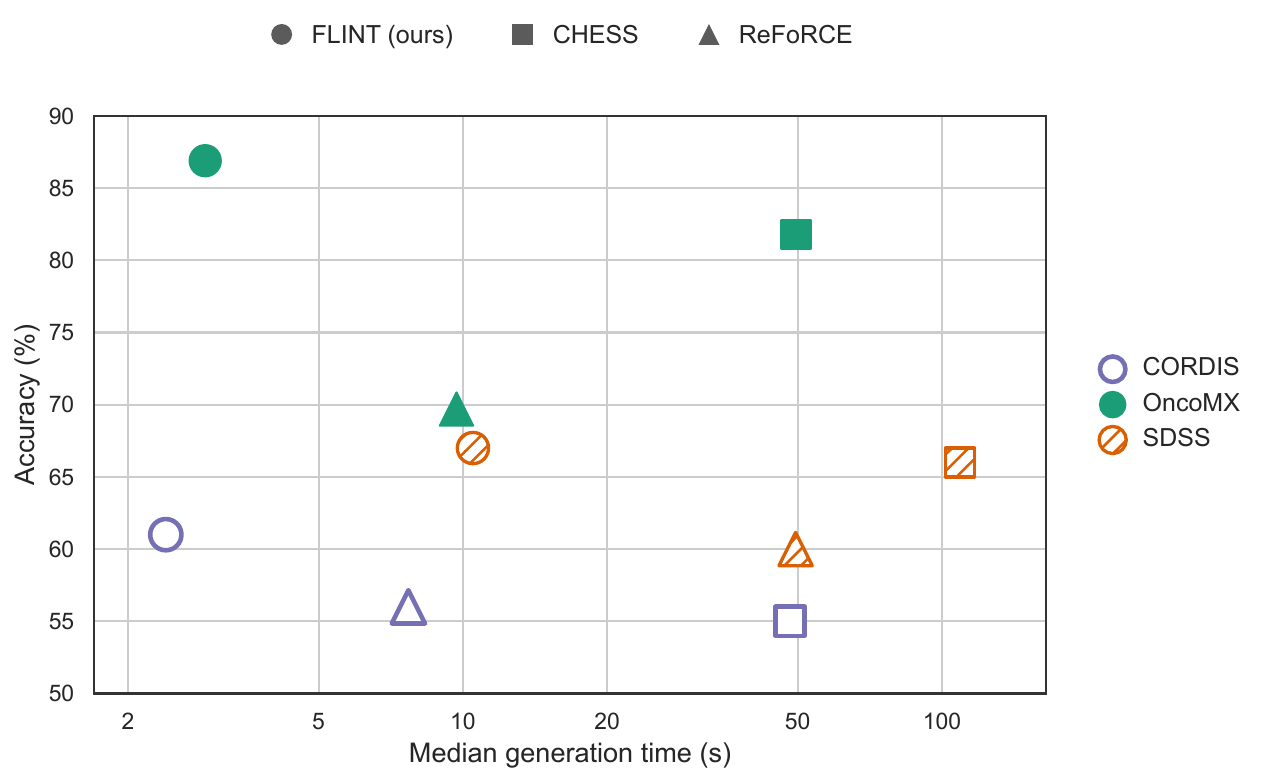}
\caption{\small{Accuracy vs.\ generation time on ScienceBenchmark (CORDIS, OncoMX, SDSS), under one harness and LLM. Top-left is best---faster and more accurate. FLINT is the most accurate on every database while generating SQL 5$\times$ faster than the multi-agent systems: CHESS issues $k{=}10$ candidate
generations per question yet trails FLINT's single pass, and ReFoRCE's iterative refinement cannot recover semantically wrong filters that still execute. Generation time excludes database execution.}}
\label{fig:other_domain}
\end{figure}

\subsection{Ablation Study}
\label{sec:ablation}

We isolate each component's contribution by disabling it while keeping the rest intact (Tab.~\ref{tab:ablation}).
For \textit{w/o Lookup}, the prompt includes static mappings for the 100 most common Financials IDs and 50 most common Transactions IDs, covering about 78\% of queries.
\textit{w/o Schema Linking}, the full schema is passed, \textit{w/o Retrieval}, no exemplar queries are provided, and \textit{w/o Reflection}, the first SQL is submitted without retry (details in App.~\ref{sec:appendix_ablations}).

\begin{table}[t]
\centering
\caption{\small{Ablation study (strict accuracy, \%). Each row disables one component while keeping the rest of the pipeline intact. Rows sorted by average degradation.}}
\small
\setlength{\tabcolsep}{5pt}
\renewcommand{\arraystretch}{1.1}
\begin{tabular}{@{}l rr rr@{}}
\toprule
& \multicolumn{2}{c}{\textbf{Financials}} & \multicolumn{2}{c}{\textbf{Transactions}} \\
\cmidrule(lr){2-3} \cmidrule(lr){4-5}
\textbf{Configuration} & \textit{Acc.} & \textit{$\Delta$} & \textit{Acc.} & \textit{$\Delta$} \\
\midrule
\rowcolor{blue!6}
\ac{FLINT} (full system) & \textbf{68.8} & --- & \textbf{65.2} & --- \\
\addlinespace[4pt]
\quad w/o Lookup Agent & 54.8 & $-$14.0 & 33.3 & $-$31.9 \\
\quad w/o Few-Shot Retrieval & 57.9 & $-$10.9 & 40.6 & $-$24.6 \\
\quad w/o Schema Linking & 65.2 & $-$3.6 & 50.7 & $-$14.5 \\
\quad w/o Reflection & 67.9 & $-$0.9 & 55.1 & $-$10.1 \\
\bottomrule
\end{tabular}
\label{tab:ablation}
\end{table}

The lookup agent is the most critical component: removing it drops accuracy by 14 points on Financials and 32 points on Transactions. Even though static mappings cover most of the queries, it cannot replace the question-specific disambiguation required to select the correct identifier. Retrieval is the second-largest contributor (11/25-point drop) despite relying on only 27 examples, indicating that diversity matters more than scale. Schema linking is especially important for Transactions (15-point drop versus 4 on Financials), where it resolves ambiguity among semantically similar columns. Reflection helps only when execution errors occur, yielding little benefit on Financials but recovering 15 of 22 failures on Transactions.

\subsection{Error Analysis}
\label{sec:error_analysis}

\ac{FLINT}'s 117 incorrect predictions fall into three main categories  (Tab~\ref{tab:error_analysis}). Column selection is the primary failure mode, accounting for 71 errors. Among partial-match cases, some return the correct values with an extra or missing non-essential column (presentation-only issues), while genuine mistakes arise from metric ambiguity (e.g., mapping ``stock price'' to adjusted close, last close, or OHLC) or from selecting an identifier column rather than its corresponding value column. 30 failures occur after the correct tables and columns have been selected, due to errors in lookup constraints. On Financials, all 15 such cases are caused by incorrect lookup resolution despite otherwise correct schema grounding. Table selection errors are uncommon (4.5\% of queries), and entity linking contributes 5 failures on Financials and 11 on Transactions. 

Table~\ref{tab:error_analysis} counts failures by strict exact-match, which scores a query wrong whenever its result table differs from the gold table. A separate manual audit finds that many of the $\geq$50\% partial-match failures return the correct values but add or omit a non-essential column. Under this interpretation, \emph{effective} accuracy rises to ${\sim}85\%$ (Financials) / ${\sim}76\%$ (Transactions), as these cases are treated as correct despite being classified as substantive errors by the strict taxonomy (App.~\ref{sec:failure_analysis}). The remainder is concentrated in value grounding, metric/column disambiguation and lookup-constraint resolution. All failure modes remain inspectable in deployment, where the analyst sees the generated SQL, the resolved IDs, and the intermediate steps.

\begin{table}[t]
\centering
\caption{\small{Error taxonomy for \ac{FLINT}'s 117 incorrect predictions. All failures produce executable SQL.}}
\small
\setlength{\tabcolsep}{5pt}
\renewcommand{\arraystretch}{1.1}
\begin{tabular}{@{}l rrr@{}}
\toprule
\textbf{Error Category} & \textbf{Fin.} & \textbf{Trans.} & \textbf{Tot.} \\
\midrule
\textit{Correct tables \& columns} & & & \\
\quad Wrong filter / aggregation & 15 & 15 & 30 \\[2pt]
\textit{Correct tables, wrong columns} & & & \\
\quad Partial match ($\geq$50\%) & 35 & 15 & 50 \\
\quad Low match ($<$50\%) & 11 & 10 & 21 \\[2pt]
\textit{Wrong table selection} & & & \\
\quad Missing / extra tables & 8 & 8 & 16 \\
\midrule
\textbf{Total} & \textbf{69} & \textbf{48} & \textbf{117} \\
\bottomrule
\end{tabular}
\label{tab:error_analysis}
\end{table}

\section{Conclusion}
\label{sec:conclusion}

We presented \acf{FLINT}, a \acs{T2S} system designed for production financial databases with deep schema, opaque integer keys as values, and questions that carry implicit domain constraints absent from the natural language. \ac{FLINT} combines schema linking, example retrieval, entity resolution, domain-specific prompt rules, and execution-based reflection into a single-pass pipeline that runs in 10--20 seconds per query. On two proprietary financial benchmarks, \ac{FLINT} outperforms the seven academic baselines. Across four additional scientific and medical domains, the same architecture generalizes with only minimal changes.
Our results suggest that the gap between academic \acs{T2S} benchmarks and production deployment is not primarily about SQL generation capability, but about the surrounding domain knowledge: knowing which tables to join, what values mean, and the intent of user questions. \ac{FLINT}'s architecture offers one reusable pattern for closing that gap.
\section*{Limitations}
\label{sec:limitations}

\ac{FLINT}'s domain rules are manually authored by subject matter experts familiar with each schema's conventions. While effective, this introduces a per-domain configuration cost that does not scale automatically to new databases without human effort. Similarly, the retrieval component requires a curated query bank. On a new schema with no historical queries, this component is unavailable and accuracy degrades. One component of our evaluation pipeline uses an LLM as judge for semantic equivalence; while we validate it against human annotations on a held-out subset, its agreement is imperfect and may introduce noise into reported accuracy. Finally, FLINT operates in a single-turn setting: It generates one SQL query per natural language question and does not support conversational refinement such as follow-up filters or corrections that reference prior context.

\bibliography{custom}

\clearpage
\appendix
\section{Dataset Examples}
\label{sec:appendix_datasets}

We present representative examples from different difficulty levels to illustrate the structural complexity of our benchmarks.
Table names are shortened for readability (e.g., \texttt{Company} for the full internal table name).

\subsection{Difficulty levels}
\label{subsec:app:difficulty}
We stratify questions by the structural complexity of the gold SQL (join count and CTE presence).
Join count is an imperfect proxy for difficulty since a 3-join query requiring implicit domain knowledge can be harder than a routine 7-join traversal. However it correlates with the number of opportunities for error accumulation and is objectively measurable.
We use three levels:

\begin{itemize}
    \item \textbf{Simple} (1--3 joins, no CTE): Entity lookups, single-table retrievals, basic filtering.
    \item \textbf{Moderate} (4--6 joins, no CTE): Standard financial data retrieval requiring the full join chain through period and metric tables.
    \item \textbf{Complex} (7+ joins or CTE): Multi-entity comparisons, cross-currency rankings, temporal aggregations.
\end{itemize}

Tab.~\ref{tab:difficulty_dist} shows the distribution.

\begin{table}[h]
\centering
\caption{\small{Question difficulty distribution by structural complexity of the gold SQL.}}
\small
\setlength{\tabcolsep}{5pt}
\renewcommand{\arraystretch}{1.25}
\begin{tabular}{@{}l rr rr@{}}
\toprule
& \multicolumn{2}{c}{\textbf{Financials}} & \multicolumn{2}{c}{\textbf{Transactions}} \\
\cmidrule(lr){2-3} \cmidrule(lr){4-5}
\textbf{Difficulty} & \textit{n} & \% & \textit{n} & \% \\
\midrule
Simple\enspace (1--3 joins) & 22 & 10 & 38 & 28 \\
Moderate\enspace (4--6 joins) & 142 & 64 & 66 & 48 \\
Complex\enspace (7{+} joins / CTE) & 57 & 26 & 34 & 25 \\
\bottomrule
\end{tabular}
\label{tab:difficulty_dist}
\end{table}

\subsection{Financials Dataset}
\label{subsec:appendix:financials_dataset}

\paragraph{Simple (1--3 joins).}
\textit{``Where is Town Pump based out of?''} \\
A geographic lookup requiring two joins to resolve state and country names from foreign key IDs.

\begin{lstlisting}[style=sql]
SELECT Company.companyName,
       State.state,
       Country.country
FROM Company
JOIN State ON Company.stateId = State.stateId
JOIN Country ON Company.countryId = Country.countryId
WHERE Company.companyId = 4222311
\end{lstlisting}

\vspace{0.5em}
\noindent\textit{2 joins. The companyId is resolved via entity linking from ``Town Pump.''}

\paragraph{Moderate (4--6 joins).}
\textit{``Nvidia revenue from 2020 to 2025''} \\
A standard financial data retrieval requiring the full six-table join chain through period types, metrics, units, and currencies.

\begin{lstlisting}[style=sql]
SELECT Company.companyName,
       FinancialData.dataItemValue AS revenue,
       DataItem.dataItemName,
       UnitType.unitTypeName AS unit,
       FinPeriod.fiscalYear,
       Currency.currencyName
FROM FinancialData
JOIN DataItem ON DataItem.dataItemId = FinancialData.dataItemId
JOIN FinPeriod ON FinPeriod.financialPeriodId
       = FinancialData.financialPeriodId
JOIN Company ON Company.companyId = FinPeriod.companyId
JOIN PeriodType ON FinPeriod.periodTypeId = PeriodType.periodTypeId
JOIN UnitType ON FinancialData.unitTypeId = UnitType.unitTypeId
JOIN Currency ON FinPeriod.currencyId = Currency.currencyId
WHERE Company.companyId = 32307
  AND FinancialData.dataItemId = 28
  AND FinPeriod.fiscalYear BETWEEN 2020 AND 2025
  AND PeriodType.periodTypeId = 1
ORDER BY FinPeriod.fiscalYear DESC
\end{lstlisting}

\vspace{0.5em}
\noindent\textit{6 joins. The dataItemId (28 = Total Revenue) and periodTypeId (1 = Annual) must be resolved from lookup tables.
This join chain is the ``backbone'' of most Financials queries.}

\paragraph{Complex (7+ joins / CTE).}
\textit{``2024 annual revenue for Alibaba''} \\
Identical intent to the previous example, but for a non-US company whose financials are reported in CNY.
A currency conversion CTE is required to normalize to USD.

\begin{lstlisting}[style=sql]
WITH CurrencyConversion AS (
  SELECT ExchangeRate.priceDate,
         ExchangeRate.currencyId,
         Currency.isoCode AS isoCodeSource,
         ER2.currencyId AS currencyIdTarget,
         C2.isoCode AS isoCodeTarget,
         ER2.priceClose / ExchangeRate.priceClose
           AS exchangeRate
  FROM ExchangeRate
  JOIN Currency ON ExchangeRate.currencyId = Currency.currencyId
  JOIN ExchangeRate ER2
    ON ExchangeRate.priceDate = ER2.priceDate
  JOIN Currency C2 ON ER2.currencyId = C2.currencyId
  WHERE ExchangeRate.latestSnapFlag = 1
    AND ER2.latestSnapFlag = 1
    AND Currency.majorCurrencyFlag = 1
    AND C2.majorCurrencyFlag = 1
)
SELECT Company.companyName,
       FinancialData.dataItemValue,
       DataItem.dataItemName,
       UnitType.unitTypeName AS unit,
       FinPeriod.fiscalYear,
       Currency.currencyName,
       FinancialData.dataItemValue
         * CurrencyConversion.exchangeRate AS revenueUSD
FROM FinancialData
JOIN DataItem ON DataItem.dataItemId = FinancialData.dataItemId
JOIN FinPeriod ON FinPeriod.financialPeriodId
       = FinancialData.financialPeriodId
JOIN Company ON Company.companyId = FinPeriod.companyId
JOIN PeriodType ON FinPeriod.periodTypeId = PeriodType.periodTypeId
JOIN UnitType ON FinancialData.unitTypeId = UnitType.unitTypeId
JOIN Currency ON FinPeriod.currencyId = Currency.currencyId
JOIN CurrencyConversion
  ON CurrencyConversion.priceDate = FinPeriod.periodEndDate
  AND CurrencyConversion.currencyId = FinPeriod.currencyId
WHERE Company.companyId = 42083601
  AND FinancialData.dataItemId = 28
  AND FinPeriod.fiscalYear = 2024
  AND PeriodType.periodTypeId = 1
  AND CurrencyConversion.isoCodeTarget = 'USD'
\end{lstlisting}

\vspace{0.5em}
\noindent\textit{10 joins (3 in CTE + 7 in main query). The CTE computes cross-rates between all major currency pairs using a triangulation through USD-denominated exchange rates. This pattern recurs in any query comparing entities across currencies.}

\subsection{Transactions Dataset}
\label{subsec:appendix:transactions_dataset}

\paragraph{Simple (1--3 joins).}
\textit{``How many IPOs were cancelled in 2020?''} \\
An aggregation query filtering on transaction type and status, both resolved via reference table IDs.

\begin{lstlisting}[style=sql]
SELECT COUNT(tr.transactionId) AS cancelledIPOs
FROM Transaction tr
JOIN TransactionType tt
  ON tr.transactionIdTypeId = tt.transactionIdTypeId
JOIN TransactionStatus ts
  ON tr.statusId = ts.statusId
WHERE tr.transactionIdTypeId = 6
  AND tr.statusId = 3
  AND tr.announcedYear = 2020
\end{lstlisting}

\vspace{0.5em}
\noindent\textit{2 joins. The transactionIdTypeId (6 = IPO) and statusId (3 = Cancelled) are resolved from reference tables.}

\paragraph{Moderate (4--6 joins).}
\textit{``When did Amazon acquire Whole Foods?''} \\
A specific acquisition lookup requiring traversal of the buyer--target relationship chain through three intermediate tables.

\begin{lstlisting}[style=sql]
SELECT target.companyName AS "Target",
       buyer.companyName AS "Buyer",
       tr.announcedDay,
       tr.announcedMonth,
       tr.announcedYear
FROM Transaction tr
JOIN TransactionType trtype
  ON tr.transactionIdTypeId = trtype.transactionIdTypeId
JOIN Company target ON tr.companyId = target.companyId
JOIN TransactionCompanyRel cr
  ON cr.transactionId = tr.transactionId
JOIN TransactionRelType crt
  ON crt.transactionToCompRelTypeId
       = cr.transactionToCompRelTypeId
JOIN CompanyRel crel ON crel.companyRelId = cr.companyRelId
JOIN Company buyer ON buyer.companyId = crel.companyId
WHERE target.companyId = 36602
  AND buyer.companyId = 18749
  AND trtype.transactionIdTypeId = 2
  AND crt.transactionToCompRelTypeId = 1
\end{lstlisting}

\vspace{0.5em}
\noindent\textit{6 joins. The four-table chain} \texttt{Transaction $\to$ TransactionCompanyRel $\to$ CompanyRel $\to$ Company} \textit{is the standard path to resolve buyer/seller relationships and recurs in most M\&A queries.}

\paragraph{Complex (7+ joins / CTE).}
\textit{``10 largest M\&A deals in the financial sector in 2022''} \\
A ranking query combining sector filtering, currency normalization, and transaction-size ordering.

\begin{lstlisting}[style=sql]
WITH CurrencyConversion AS (
  SELECT ExchangeRate.priceDate,
         ExchangeRate.currencyId,
         Currency.isoCode AS isoCodeSource,
         ER2.currencyId AS currencyIdTarget,
         C2.isoCode AS isoCodeTarget,
         ER2.priceClose / ExchangeRate.priceClose
           AS exchangeRate
  FROM ExchangeRate
  JOIN Currency
    ON ExchangeRate.currencyId = Currency.currencyId
  JOIN ExchangeRate ER2
    ON ExchangeRate.priceDate = ER2.priceDate
  JOIN Currency C2 ON ER2.currencyId = C2.currencyId
  WHERE ExchangeRate.latestSnapFlag = 1
    AND ER2.latestSnapFlag = 1
    AND C2.majorCurrencyFlag = 1
)
SELECT tr.transactionId,
       Company.companyName,
       tr.announcedYear,
       TransactionType.transactionIdTypeName,
       Industry.simpleIndustryDescription,
       tr.transactionSize
         * exrate.exchangeRate AS transactionSizeUSD
FROM Transaction tr
JOIN TransactionType
  ON tr.transactionIdTypeId
       = TransactionType.transactionIdTypeId
JOIN Company ON tr.companyId = Company.companyId
JOIN Industry
  ON Company.simpleIndustryId = Industry.simpleIndustryId
JOIN CurrencyConversion exrate
  ON tr.currencyId = exrate.currencyId
  AND DATE_FROM_PARTS(
        tr.announcedYear, tr.announcedMonth,
        tr.announcedDay) = exrate.priceDate
WHERE Industry.simpleIndustryId IN (46,48,49,50,51)
  AND TransactionType.transactionIdTypeId = 2
  AND tr.announcedYear = 2022
  AND tr.transactionSize > 0
  AND exrate.isoCodeTarget = 'USD'
ORDER BY transactionSizeUSD DESC
LIMIT 10
\end{lstlisting}

\vspace{0.5em}
\noindent\textit{7 joins (3 in CTE + 4 in main query). Combines the currency conversion CTE with industry filtering. The} \texttt{IN (46,48,49,50,51)} \textit{clause reflects the multiple sub-industry IDs that map to ``Financial Sector'', another reference table resolution challenge.}

\subsection{Key Observations}

These examples illustrate three properties that make our benchmarks challenging for general-purpose \acs{T2S} systems:

\begin{enumerate}
    \item \textbf{Deep normalization}: Even simple questions require resolving opaque integer IDs (e.g., \texttt{dataItemId = 28} for revenue, \texttt{periodTypeId = 1} for annual) from lookup tables. Without a dedicated lookup mechanism, models cannot produce correct filter predicates (Fig.~\ref{fig:opaque_ids}).

    \item \textbf{Long join chains}: The standard financial data retrieval pattern requires 6 joins as a baseline. This is the ``moderate'' difficulty, i.e. the minimum join path for accessing any time-series financial metric.

    \item \textbf{Currency normalization}: Any cross-country comparison requires a CTE that triangulates exchange rates, adding 3 joins and substantial SQL complexity. This pattern cannot be discovered through schema inspection alone. It requires domain knowledge about how the exchange rate table encodes USD-denominated rates.
\end{enumerate}

 \begin{figure}[t]
  \centering
  \small

  \begin{tcolorbox}[colback=gray!5, colframe=gray!60, title={\small\textbf{Schema declaration (visible to the model)}}, fonttitle=\small, boxrule=0.4pt, arc=2pt]
  \ttfamily
  ciqFinancialData.dataItemId INTEGER REFERENCES ciqDataItem(dataItemId)\\
  ciqFinancialData.periodTypeId INTEGER REFERENCES ciqPeriodType(periodTypeId)
  \end{tcolorbox}
  
  \vspace{4pt}
  {\centering\small\textit{Question: ``What was the quarterly EBITDA?''}\par}
  \vspace{2pt}
  {\centering $\Downarrow$ \textsc{Lookup Agent queries reference tables} $\Downarrow$\par}
  \vspace{4pt}

  \begin{tcolorbox}[colback=blue!3, colframe=blue!40, title={\small\textbf{Reference tables}}, fonttitle=\small, boxrule=0.4pt, arc=2pt, halign=center]
  \centering
  \begin{tabular}{@{}rl@{\hskip 24pt}rl@{}}
  \multicolumn{2}{c}{\texttt{ciqPeriodType}} & \multicolumn{2}{c}{\texttt{ciqDataItem}} \\[2pt]
  \texttt{1} & Annual & \texttt{28} & Total Revenue \\
  \rowcolor{yellow!20}
  \texttt{2} & Quarterly & \texttt{42} & EBITDA \\
  \texttt{3} & LTM & \texttt{105} & Net Income \\
  \end{tabular}
  \end{tcolorbox}

  \vspace{2pt}
  {\centering $\Downarrow$ \small\textrm{Resolved: \texttt{periodTypeId=2 AND dataItemId=42}}\par}

  \caption{Opaque identifiers in Financials. The schema exposes only integer foreign keys. As a result, resolving ``quarterly EBITDA'' to concrete filter values requires querying reference tables, a step the Lookup Agent performs automatically.}
  \label{fig:opaque_ids}
  \end{figure}

\subsection{Benchmark Size}

Our benchmark (359 questions) is smaller than Spider (1{,}034 dev) or BIRD (1{,}534 dev).
We argue this is sufficient for two reasons.
First, both datasets share a single fixed schema, eliminating the cross-database variance that inflates sample size requirements in multi-schema benchmarks.
Second, a 95\% Clopper-Pearson confidence interval at 77\% accuracy with $n{=}221$ is $\pm$5.5 points, narrow enough to distinguish \ac{FLINT} from all baselines at $p < 0.001$.
The benchmark is not intended to estimate fine-grained performance differences between systems within a few points of each other; rather, it is designed to measure whether domain-specialized infrastructure closes the gap that general-purpose methods cannot.

\subsection{Comparison with Existing \acs{T2S} Benchmarks}
\label{sec:benchmark_comparison}

Table~\ref{tab:benchmark_comparison} contrasts our benchmarks with Spider~\cite{yu-etal-2018-spider} and BIRD~\cite{li2024bird}.

\begin{table}[h]
\centering
\caption{\small{Comparison with Spider and BIRD. Key differentiators: deep normalization with opaque IDs, mandatory multi-table joins for even simple queries, and production-deployed schemas.}}
\small
\setlength{\tabcolsep}{4pt}
\renewcommand{\arraystretch}{1.25}
\begin{tabular}{@{}l ccc@{}}
\toprule
\textbf{Property} & \textbf{Spider} & \textbf{BIRD} & \textbf{Ours} \\
\midrule
Avg.\ joins per query & 1.2 & 1.8 & 5.6 \\
Max joins (single query) & 4 & 6 & 13 \\
Reference/lookup tables & 0 & 0--2 & 12--16 \\
Opaque integer IDs & \xmark & \xmark & \cmark \\
Currency conversion & \xmark & \xmark & \cmark \\
Self-describing values & \cmark & \cmark & \xmark \\
Schema size (tables) & 4--8 & 5--15 & 18--24 \\
Production deployment & \xmark & \xmark & \cmark \\
\bottomrule
\end{tabular}
\label{tab:benchmark_comparison}
\end{table}

Three properties distinguish our benchmarks:

\paragraph{Opaque schemas.}
Spider and BIRD filter on human-readable values (\texttt{country = `USA'}).
Our schemas encode concepts as integer foreign keys (\texttt{dataItemId = 28} for ``Total Revenue''), requiring a lookup step that no existing benchmark demands.

\paragraph{Structural depth as baseline.}
On Spider, 78\% of queries use 0 to 1 joins.
On our benchmarks, most of our queries use 4 -- 6 joins. The standard path traverses \texttt{Company $\to$ FinPeriod $\to$ FinancialData $\to$ DataItem $\to$ PeriodType $\to$ UnitType $\to$ Currency}.
BIRD increases schema scale modestly but not join depth, as about 84\% of the queries have at most 1 join.

\paragraph{Implicit domain conventions.}
Questions carry constraints absent from the natural language. For example, ``stock price'' implies \texttt{priceClose}; ``revenue'' implies annual (\texttt{periodTypeId = 1}); ``market cap'' requires \texttt{primaryFlag = 1}; and cross-country comparisons require currency normalization.
Spider and BIRD questions are self-contained, while ours require domain knowledge to disambiguate.

\section{Configuration Cost}
\label{sec:appendix_config}

The bulk of configuration effort is authoring the query bank: relevant questions, their gold SQL, and a reasoning trace. Each takes longer than a routine evaluation question (which averages ${\sim}30$ minutes including 2--3 review rounds) because of the chain-of-thought walkthrough, roughly 12 person-hours for Transactions and 18 person-hours for Financials. The schema CSV is usually straightforward, since production databases already carry metadata or customer-facing documentation. Instruction prompts and example traces
add ${\sim}4$--$8$ person-hours for a complex schema.

All in, a moderate-complexity dataset takes roughly a person-week, while a complex one (24 tables, 16 reference tables, multiple implicit conventions) two to three weeks including iterative refinement as production traffic surfaces gaps. Once configured, adding an example for a novel pattern is ${\sim}30$ minutes.

Sensitivity is asymmetric across both artifacts and datasets. Domain rules are a prerequisite. The rules alone can beat various multi-agent systems. Returns are not linear in bank size, beyond a small,
diverse set, tightening a rule tends to help more than adding examples. 
\section{Evaluation Pipeline Details}
\label{sec:appendix_ast}

Evaluating \acs{T2S} correctness is fundamentally harder than string matching: Syntactically different queries can produce identical results, while structurally similar queries can return different data due to a single wrong filter value.
Standard benchmarks sidestep this with execution accuracy (run both queries, compare outputs), but this is insufficient for production-scale schemas where result-set equality is sensitive to row ordering, null handling, and floating-point precision.

\subsection{Comparison with Standard Benchmarks}

Spider~\cite{yu-etal-2018-spider} uses exact set match (ESM): The gold and predicted SQL are decomposed into components (\texttt{SELECT}, \texttt{WHERE}, \texttt{GROUP BY}, \texttt{ORDER BY}) and compared structurally.
ESM is brittle, it rejects semantically equivalent queries with different syntactic structure (e.g., a subquery reformulated as a \texttt{JOIN}, or an \texttt{EXISTS} replaced by \texttt{IN}).
BIRD~\cite{li2024bird} improves on this by using execution accuracy (EX): Both queries run against the database and the result sets are compared.
However, EX treats any difference in output as failure, even when differences arise from column ordering, alias naming, or equivalent NULL handling.

Our evaluation addresses both limitations.
We do not rely on structural matching alone (unlike ESM), since our queries have many valid reformulations (CTEs vs.\ subqueries, different join orderings, aliasing choices).
We go beyond naive execution comparison (unlike EX) by incorporating column alignment, order-invariant comparison, AST canonicalization, and an LLM judge for semantic equivalence.

\subsection{Three-Tier Pipeline}

A single evaluation method is insufficient for normalized financial schemas.
Consider a query retrieving Apple's annual revenue: the gold SQL joins through \texttt{ciqCompany}, \texttt{ciqSecurity}, \texttt{ciqTradingItem}, \texttt{ciqFinPeriod}, \texttt{ciqFinancialData}, and \texttt{ciqDataItem} with specific filter predicates.
A correct prediction might use different aliases, reorder joins, wrap part of the logic in a CTE, or select columns in a different order, all the while producing identical results.

\begin{scriptsize}
\begin{verbatim}
-- Gold: two aliases of ciqCompany via inline joins
SELECT target.companyName, acquirer.companyName
FROM ciqTransactionV2 tr
JOIN ciqCompany target ON tr.companyId = target.companyId
JOIN ciqTransactionToCompanyRelV2 cr ON ...
JOIN ciqCompany acquirer ON crel.companyId = acquirer.companyId
WHERE target.companyId = 24937

-- Generated: equivalent result, 
-- CTE + different join topology
WITH TargetCo AS 
    (SELECT companyId FROM ciqCompany WHERE companyId = 24937)
SELECT c.companyName, acq.companyName
FROM ciqTransactionV2 tr
JOIN TargetCo tc ON tr.companyId = tc.companyId
JOIN ciqCompany c ON tr.companyId = c.companyId
JOIN ciqCompany acq ON crel.companyId = acq.companyId ...
\end{verbatim}
\end{scriptsize}

Both return identical rows, but the CTE node and rearranged self-join aliases produce an irreducible tree difference.
No single comparison method handles all such cases, motivating a cascading pipeline where each tier catches what the previous one misses.

The first tier parses both gold and generated queries into abstract syntax trees using \texttt{sqlglot} and applies canonicalizing transformations: alias normalization (replacing all aliases with canonical names derived from the underlying table), date function standardization, \texttt{JOIN} reordering (sorting inner joins by table name), and predicate normalization (canonicalizing commutative predicates and \texttt{IN}-clause value ordering).
After canonicalization, tree edit distance determines whether the remaining differences are semantically meaningful.
For self-joins, all alias permutations are tested, tractable for our schema's typical 2--3 self-join aliases.
Queries that are structurally equivalent modulo these transformations are matched without execution.

When AST matching fails, the second tier executes both queries against the production database and compares outputs.
Execution failures score 0 immediately.
For successful queries, gold columns are matched to generated columns via \texttt{select} clauses from SQLs or case-insensitive name matching, falling back to data-type and value-distribution similarity for ambiguous cases.
Values are compared with order-invariant set equality, and numeric comparisons apply a relative tolerance of $10^{-6}$ for floating-point differences.

When output tables still differ, the third tier invokes an LLM judge that receives the original question, gold and generated results. The judge determines semantic equivalence in context, handling cases where column names differ but values match, results are formatted differently. We manually validated 50 randomly sampled judge decisions and found 94\% agreement with human annotators (47/50 correct, 2 false positives, 1 false negative).

\section{Baseline Adaptation Details}
\label{sec:appendix_baselines}

All baseline methods were originally evaluated on Spider~\cite{yu-etal-2018-spider} or BIRD~\cite{li2024bird}, which feature flat schemas with 1--3 joins per query.
Adapting them to our heavily-normalized financial databases (6+ joins on average, 16+ reference tables) required both infrastructure decisions and architectural choices.
This appendix documents the adaptations made and the rationale behind each.

\subsection{Shared Infrastructure}
\label{subsec:appendix_baselines-shared_infra}

All baselines receive the same infrastructure that does not constitute a methodological advantage:

\begin{itemize}
    \item \textbf{Database connection:} Read-only Snowflake access to execute SQL against the production schema.
    \item \textbf{Entity linking:} Company name disambiguation via an external service (Kensho Link) that resolves natural-language entity references (e.g., ``Nvidia'' $\to$ \texttt{companyId=32307}). This is provided to methods that operate within a pipeline where entity resolution is a separate preprocessing step. Methods that instead explore the database directly (MAC-SQL) achieve equivalent resolution through interactive queries.
    \item \textbf{CTE injection:} Gold SQL queries reference materialized CTEs for currency conversion and temporal normalization. These CTEs are injected into both gold and predicted queries at evaluation time, they are infrastructure specific rather than model output.
    \item \textbf{Post-processing:} All baselines use the same safety validator and temporal postprocessor as \ac{FLINT} (Section~\ref{sec:generation}).
    \item \textbf{Evaluation:} Three-tier evaluation (heuristic table comparison, SQL AST matching, LLM judge) applied uniformly to all methods.
\end{itemize}

\subsection{Direct Prompting (Zero-Shot and Few-Shot)}

To establish a floor for what the LLM can achieve without any \acs{T2S}-specific architecture, we evaluate two direct prompting configurations:

\begin{itemize}
    \item \textbf{Zero-shot:} The full schema (all tables and columns with descriptions) and entity IDs from the external linking service are provided as prompt context. The model generates SQL in a single pass with no examples, no lookup, no schema linking, and no reflection.
    \item \textbf{Few-shot ($k$=3):} Same as zero-shot, plus $k{=}3$ similar examples retrieved from the query bank (question and SQL only, no intermediate reasoning traces).
\end{itemize}

Both configurations use the same entity linking service as \ac{FLINT}, ensuring company names are resolved to correct integer IDs.
The gap between few-shot and \ac{FLINT} isolates the contribution of the domain-specialized pipeline (lookup agent, schema linking, domain-specific prompts, reflection).

\subsection{CHESS}

CHESS~\cite{talaei2024chess} is a multi-agent \acs{T2S} system that achieves state-of-the-art accuracy on BIRD through schema linking, candidate generation with self-consistency, and unit-test-based selection.

We use the original CHESS codebase with minimal modification:
\begin{itemize}
    \item Replace SQLite execution with Snowflake connection.
    \item Provide schema documentation as column-level descriptions (matching the format CHESS expects from BIRD's \texttt{database\_description} files).
    \item Entity linking is \emph{not} provided. CHESS handles entity resolution through its own schema-value matching and column-value retrieval modules.
    \item Evaluate in both single-candidate (k=1) and full self-consistency (k=10 candidates with unit-test selection) configurations.
\end{itemize}
All core CHESS components are preserved; the only change is the execution backend (SQLite $\to$ Snowflake).

\subsection{DAIL-SQL}

DAIL-SQL~\cite{gao2024dail} introduces question-SQL example selection via token-level similarity and a structured prompt format that separates schema representation from in-context examples.

DAIL-SQL is a prompt-engineering method without its own pipeline infrastructure. We implement it within our pipeline, providing:
\begin{itemize}
    \item Schema linking via embedding-based retrieval to identify relevant tables.
    \item Lookup agent for reference-table constraint resolution (shared infrastructure).
    \item Entity linking for company name disambiguation (shared infrastructure).
    \item Similar query retrieval from a query bank (the DAIL-SQL ``example selection'' step).
    \item DAIL-style prompts: generic schema linking instructions and SQL generation prompts without domain-specific rules (no currency conversion hints, no period-type defaults, no metric disambiguation guidance).
    \item No reflection (zero retry attempts).
    \item No structural validation.
\end{itemize}
We preserve DAIL-SQL's prompt-level contribution (generic prompts + example selection) while providing the infrastructure necessary to operate on our schema, without entity linking, no method can resolve company names against a 20 million-company table.
This gives DAIL-SQL a stronger starting point than it would have with its original prompts alone, making our comparison conservative.

\subsection{DIN-SQL}

DIN-SQL~\cite{pourreza2024din} decomposes \acs{T2S} into sub-tasks: schema linking, query classification (easy/medium/hard), and difficulty-aware SQL generation with self-correction.

Like DAIL-SQL, DIN-SQL is a prompt-engineering approach. We implement it within our pipeline:
\begin{itemize}
    \item Same shared infrastructure as DAIL-SQL (entity linking, lookup agent, similar query retrieval).
    \item DIN-SQL-style prompts: step-by-step schema linking instructions and classification-aware generation prompts.
    \item Self-correction with 1 reflection attempt (matching the DIN-SQL paper's self-correction step).
    \item No structural validation.
\end{itemize}
We preserve DIN-SQL's key contribution: decomposition into classification-aware sub-tasks with self-correction through the prompt structure and single reflection attempt.
The shared infrastructure gives DIN-SQL the same advantages as DAIL-SQL, making this a conservative comparison.

\subsection{MAC-SQL}

MAC-SQL~\cite{wang2024macsql} proposes a Selector $\to$ Decomposer $\to$ Refiner architecture where the Decomposer has interactive database access through a multi-turn tool-calling loop.

We implement MAC-SQL as a standalone pipeline, independent of \ac{FLINT}'s infrastructure:
\begin{itemize}
    \item \textbf{Selector:} Single LLM call that receives the full schema (all tables and columns with descriptions) and returns a pruned subset relevant to the question. No embedding-based retrieval, only pure LLM schema selection.
    \item \textbf{Decomposer:} Multi-turn agentic loop with two tools: \texttt{execute\_sql} (run intermediate queries against Snowflake, returns truncated results) and \texttt{submit\_final\_sql} (terminate with final answer). Budget of 5 intermediate queries. The model explores reference tables, verifies column values, and tests subqueries before submitting.
    \item \textbf{Refiner:} If the final SQL fails execution, a single-shot LLM call attempts to fix it given the error message. Maximum 2 retry attempts.
    \item No entity linking, no lookup agent, no similar query retrieval, no schema linking embeddings.
\end{itemize}

The Decomposer's tool-calling loop serves as a functional equivalent to entity linking. Instead of resolving entities via an external service, the model queries the database directly (e.g., \texttt{SELECT companyId FROM ciqCompany WHERE companyName ILIKE '\%Nvidia\%'}).
This means MAC-SQL does not require Kensho Link infrastructure, achieving entity resolution through interactive exploration.

\subsection{Chase-SQL}

Chase-SQL~\cite{pourreza2025chase} generates multiple candidate SQL queries using diverse reasoning strategies and selects the best one through pairwise comparison.

We implement Chase-SQL as a standalone pipeline:
\begin{itemize}
    \item \textbf{Three generators} (run in parallel):
    \begin{enumerate}
        \item \emph{Divide \& Conquer CoT:} Decomposes the question into sub-problems, solves each, combines via CTEs.
        \item \emph{Query Plan CoT:} Reasons through execution plan steps (scan $\to$ filter $\to$ join $\to$ aggregate $\to$ project).
        \item \emph{Online Synthetic:} Generates 3 similar question-SQL examples on the fly, then uses them as few-shot context.
    \end{enumerate}
    \item \textbf{Query Fixer:} Failed candidates receive up to 2 retry attempts with the error message.
    \item \textbf{Pairwise Selection:} All executable candidates are compared in a tournament. Each pair is evaluated by an LLM judge that sees both queries, their results (first rows), and the original question. The candidate with the most wins is selected.
    \item All generators receive the full schema as text. No schema linking, no entity linking, no lookup agent.
\end{itemize}
We preserve all three generation strategies and the pairwise selection mechanism from the paper.
Unlike MAC-SQL, Chase-SQL has no mechanism for interactive database exploration or entity resolution, instead generating SQL in a single pass, making it more vulnerable to reference-table resolution failures.

\subsection{FlexSQL}

FlexSQL proposes a five-stage pipeline: schema exploration, plan generation, plan clarification, SQL generation, and self-evaluation with retry.
We implement all five stages with entity linking and Snowflake execution.
In the 3-shot configuration, retrieved examples (with schema links and ID resolutions) are injected at plan generation to inform table selection and join paths.
Self-evaluation re-executes and retries once on failure.

\subsection{ReFoRCE}

ReFoRCE~\cite{reforce} uses iterative self-refinement: An exploration phase generates diagnostic queries against the database, then the model iteratively refines SQL using execution feedback (up to $N{=}5$ iterations, early stop on success).
We implement the full exploration and refinement loop with Snowflake execution.
In the 3-shot configuration, retrieved examples with ID resolutions are injected into the initial generation prompt.

\subsection{Summary of Adaptations}
\label{subsec:appendix_baselines-summary_adaptations}

Table~\ref{tab:adaptation_summary} summarizes which infrastructure components each method receives.

\begin{table*}[h]
\centering
\caption{\small{Infrastructure and architectural components per method. $^*$CHESS uses its own schema-value matching. $^\dagger$MAC-SQL achieves equivalent resolution through interactive DB queries.}}
\small
\setlength{\tabcolsep}{4pt}
\renewcommand{\arraystretch}{1.3}
\begin{tabular}{@{}l ccccccc@{}}
\toprule
\textbf{Component} & \rotatebox{55}{\textbf{\ac{FLINT}}} & \rotatebox{55}{\textbf{CHESS}} & \rotatebox{55}{\textbf{DAIL/DIN}} & \rotatebox{55}{\textbf{MAC-SQL}} & \rotatebox{55}{\textbf{Chase-SQL}} & \rotatebox{55}{\textbf{FlexSQL}} & \rotatebox{55}{\textbf{ReFoRCE}} \\
\midrule
Entity linking & \cmark & \xmark$^*$ & \cmark & \xmark$^\dagger$ & \xmark & \cmark & \xmark \\
Lookup agent & \cmark & \xmark & \cmark & \xmark$^\dagger$ & \xmark & \xmark & \xmark \\
Schema linking & \cmark & \cmark & \cmark & LLM & \xmark & \cmark & \xmark \\
Similar query retrieval & \cmark & \xmark & \cmark & \xmark & \xmark & \cmark & \cmark \\
Reflection / retry & \cmark & \xmark & DIN & Refiner & Fixer & \cmark & \cmark \\
DB exploration & \xmark & \xmark & \xmark & \cmark & \xmark & \xmark & \cmark \\
Multi-candidate & \xmark & \cmark & \xmark & \xmark & \cmark & \xmark & \xmark \\
\bottomrule
\end{tabular}

\label{tab:adaptation_summary}
\end{table*}

The shared infrastructure (entity linking, lookup agent) given to DAIL-SQL, DIN-SQL, and FlexSQL makes our comparison \emph{conservative}. Without it, these methods cannot resolve company names or reference-table IDs at all, and would achieve near-zero accuracy.
We chose to give them the strongest possible implementation to demonstrate that the gap with \ac{FLINT} is due to the generation strategy and domain-specific prompt engineering, not missing pre-processing.
MAC-SQL, Chase-SQL, and ReFoRCE receive no shared infrastructure because their architectures are self-contained. MAC-SQL and ReFoRCE achieve entity resolution through database exploration, while Chase-SQL's lack of any resolution mechanism is an inherent architectural limitation.

We acknowledge several tensions in this evaluation:
\begin{enumerate}
    \item \textbf{DAIL-SQL and DIN-SQL within our pipeline:} By providing entity linking and lookup, we are effectively comparing ``\ac{FLINT} with generic prompts'' against ``\ac{FLINT} with domain-specific prompts.'' A literal port of these methods (standalone, no infrastructure) yields near-zero accuracy and provides no useful signal. We believe that the current comparison better isolates the contribution of prompt engineering.
    \item \textbf{Chase-SQL without value retrieval:} The original Chase-SQL uses LSH-based value retrieval to inject relevant column values into generator prompts. Our implementation omits this component, which may disadvantage Chase-SQL on entity-resolution-heavy queries. However, adapting LSH retrieval to our reference-table structure (where values are IDs rather than human-readable strings) would require non-trivial re-engineering that goes beyond the paper's design.
    \item \textbf{MAC-SQL Selector:} The original MAC-SQL uses embedding-based schema selection, while our implementation uses a single LLM call over the full schema text. On schemas of our size (18--24 tables), the LLM-based Selector may perform differently than embedding retrieval, though both achieve the same goal of schema pruning.
\end{enumerate}

\subsection{Failure Analysis}
\label{sec:failure_analysis}

Strong baseline methods achieve 80--90\% on Spider and 60--75\% on BIRD, where column values are human-readable and queries require 1--2 joins.
On our schemas, the shared failure mode is \emph{inability to resolve opaque reference-table IDs}.

To quantify this directly, we define \textbf{filter predicate accuracy}: For each query requiring numeric ID filters (95\% of queries), we check whether all gold filter values appear in the generated SQL.
Table~\ref{tab:id_resolution} reports this metric alongside the fraction of failures attributable to wrong IDs.

\begin{table}[h]
\centering
\caption{\small{Filter predicate analysis. \textbf{ID Acc.}: fraction of queries where all required numeric IDs are correctly resolved. \textbf{Fail$\to$ID}: among incorrect predictions, the fraction where $\geq$1 required ID is wrong. On Financials, 64--96\% of all failures are attributable to incorrect filter values.}}
\small
\setlength{\tabcolsep}{5pt}
\renewcommand{\arraystretch}{1.3}
\begin{tabular}{@{}l rr rr@{}}
\toprule
& \multicolumn{2}{c}{\textbf{Financials}} & \multicolumn{2}{c}{\textbf{Transactions}} \\
\cmidrule(lr){2-3} \cmidrule(lr){4-5}
\textbf{Method} & \textit{ID Acc.} & \textit{Fail{\small$\to$}ID} & \textit{ID Acc.} & \textit{Fail{\small$\to$}ID} \\
\midrule
\rowcolor{blue!6}
\ac{FLINT} (full) & \textbf{94\%} & 13\% & \textbf{75\%} & 28\% \\
\quad w/o Lookup Agent & 10\% & 93\% & 7\% & 94\% \\
\addlinespace[4pt]
CHESS ($k{=}10$) & 46\% & 64\% & 58\% & 61\% \\
MAC-SQL & 30\% & 76\% & 39\% & 62\% \\
DAIL-SQL & 21\% & 83\% & 28\% & 79\% \\
DIN-SQL & 22\% & 85\% & 29\% & 76\% \\
FlexSQL & 44\% & 76\% & 69\% & 46\% \\
ReFoRCE & 24\% & 82\% & 28\% & 79\% \\
Chase-SQL & 6\% & 96\% & 7\% & 94\% \\
\bottomrule
\end{tabular}

\label{tab:id_resolution}
\end{table}

\ac{FLINT} resolves 94\%/75\% of required IDs (Financials/Transactions); without its lookup agent, ID accuracy drops to 10\%/7\%, identical to Chase-SQL and 93--94\% of failures become ID errors.
Among baselines, Chase-SQL resolves all required IDs in only 6--7\% of queries.
DAIL-SQL and DIN-SQL receive both entity linking \emph{and} the lookup agent, yet resolve IDs in only 21--29\% of queries.
This reveals that providing correct ID candidates is insufficient, domain-specific prompts are required to select the correct value among ambiguous options (e.g., distinguishing Total Revenue from Net Revenue, or choosing the correct period type).
The gap between \ac{FLINT} and DAIL/DIN is not infrastructure access but prompt-level domain knowledge that guides ID disambiguation.

\begin{itemize}
    \item \textbf{CHESS:} Self-consistency cannot fix ID errors consistent across samples. All $k$ candidates share the same incorrect filter predicates. Its schema-value matching provides partial resolution (46\% ID accuracy), explaining its relative strength.
    \item \textbf{Chase-SQL:} Three generation strategies converge to identical SQL when the bottleneck is what values to filter on, not how to structure the query. Without any value-retrieval mechanism, ID resolution is near-zero.
    \item \textbf{MAC-SQL:} Interactive access allows discovering some reference values (30--39\% ID accuracy), but its 5-query budget is exhausted before resolving all required lookups in complex queries.
    \item \textbf{ReFoRCE:} Refinement fixes execution errors but cannot fix semantically wrong filter values, the query executes successfully with wrong IDs, so no error signal triggers refinement.
    \item \textbf{DAIL-SQL / DIN-SQL:} These receive both entity linking and the lookup agent, yet achieve only 21--29\% ID accuracy. The lookup agent provides candidate IDs, but generic prompts lack the domain-specific rules needed to disambiguate among similar options (e.g., which revenue metric, which period type).
    \item \textbf{FlexSQL:} FlexSQL on Transactions is an exception where less than half of the errors are attributable to the ID resolution. On Transactions, FlexSQL gets the right tables only 65\% of the time and right columns 46\% which explains the bulk of non-ID failures. It has good ID resolution (69\%) but picks wrong tables/columns, meaning its plan-based schema exploration still misses structural paths. 
\end{itemize}

The ablation confirms this diagnosis: removing \ac{FLINT}'s lookup agent reduces accuracy to 55\%/33\% (Financials/Transactions), comparable to the strongest baselines. The cleanest test holds the schema fixed: given \ac{FLINT}'s own schema links, DAIL/DIN
retain ${\sim}0.97$ table and 0.70--0.81 column overlap with gold, yet still score only 32--40\%, so their remaining errors are wrong filter IDs, not wrong tables or columns, a value-grounding gap. Methods that prune aggressively instead lose the schema itself:
MAC-SQL keeps only 0.39/0.20 of gold columns (Fin/Trans), and ReFoRCE's iterative refinement prunes hard enough that overlap collapses, a distinct failure mode that Table~\ref{tab:id_resolution} does not separate. Once the correct schema is supplied, value grounding is what remains.

\paragraph{Understanding FLINT Failures.}
Finally inspecting \ac{FLINT}'s own Transactions schema-linking failures, confusion concentrates in two column groups. (1) The \emph{buyer/target relationship chain}: the intermediate transaction-to-company and company-relationship tables with their parallel party-ID columns and relationship-type key (the multi-hop who-transacted-with-whom path) is dropped without pruning. (2) \emph{Currency/status
scaffolding}: the exchange-rate/currency-conversion tables and their flag columns, plus the transaction-status table is over-selected. This suggests Transactions-like schemas benefit from more explicit join-path and currency-usage guidance in the schema-linking prompt.

Table~\ref{tab:accuracy_by_difficulty} shows accuracy of \ac{FLINT} stratified by structural complexity. Performance peaks on moderate queries (75\%/70\%), the system's design point, where the standard join chain is well-covered by few-shot examples and lookup constraints. Simple queries (64\%/66\%) perform \emph{worse} than moderate despite lower structural complexity. So we conduct a separate manual audit to understand this phenomenon. Crediting
only failures that return the correct values while differing in a non-essential column/table raises \emph{effective}
accuracy from $68.8\%/65.2\%$ to ${\sim}85\%/{\sim}76\%$. We call this a presentation-only failure. The residual simple-query
gap is not under-specification: of the eight simple-query failures on Financials, about half are presentation-only, and the genuine ones are entity-linking mistakes on otherwise-trivial questions and a tendency to \emph{over-engineer} them (adding spurious currency conversion or extra joins). This over-engineering is the flip side
of the robustness that helps on complex queries. Complex queries (54\%/52\%) degrade from compounding errors across 7+ joins.

\begin{table}[t]
\centering
\caption{\small{\ac{FLINT} accuracy (\%) by query difficulty.}}
\small
\setlength{\tabcolsep}{5pt}
\renewcommand{\arraystretch}{1.3}
\begin{tabular}{@{}l cc cc@{}}
\toprule
& \multicolumn{2}{c}{\textbf{Financials}} & \multicolumn{2}{c}{\textbf{Transactions}} \\
\cmidrule(lr){2-3} \cmidrule(lr){4-5}
\textbf{Difficulty} & \textit{n}/\textit{N} & Acc. & \textit{n}/\textit{N} & Acc. \\
\midrule
Simple\enspace (1--3 joins) & 14/22 & 64 & 25/38 & 66 \\
\rowcolor{green!6}
Moderate\enspace (4--6 joins) & 107/142 & \textbf{75} & 47/66 & \textbf{71} \\
Complex\enspace (7{+} / CTE) & 31/57 & 54 & 18/34 & 53 \\[3pt]
\midrule
\rowcolor{blue!6}
\textbf{Overall} & \textbf{152/221} & \textbf{69} & \textbf{90/138} & \textbf{65} \\
\bottomrule
\end{tabular}
\label{tab:accuracy_by_difficulty}
\end{table}

\subsection{Latency Analysis}
\label{sec:latency_analysis}

\begin{figure*}[h]
\centering
\includegraphics[width=\linewidth]{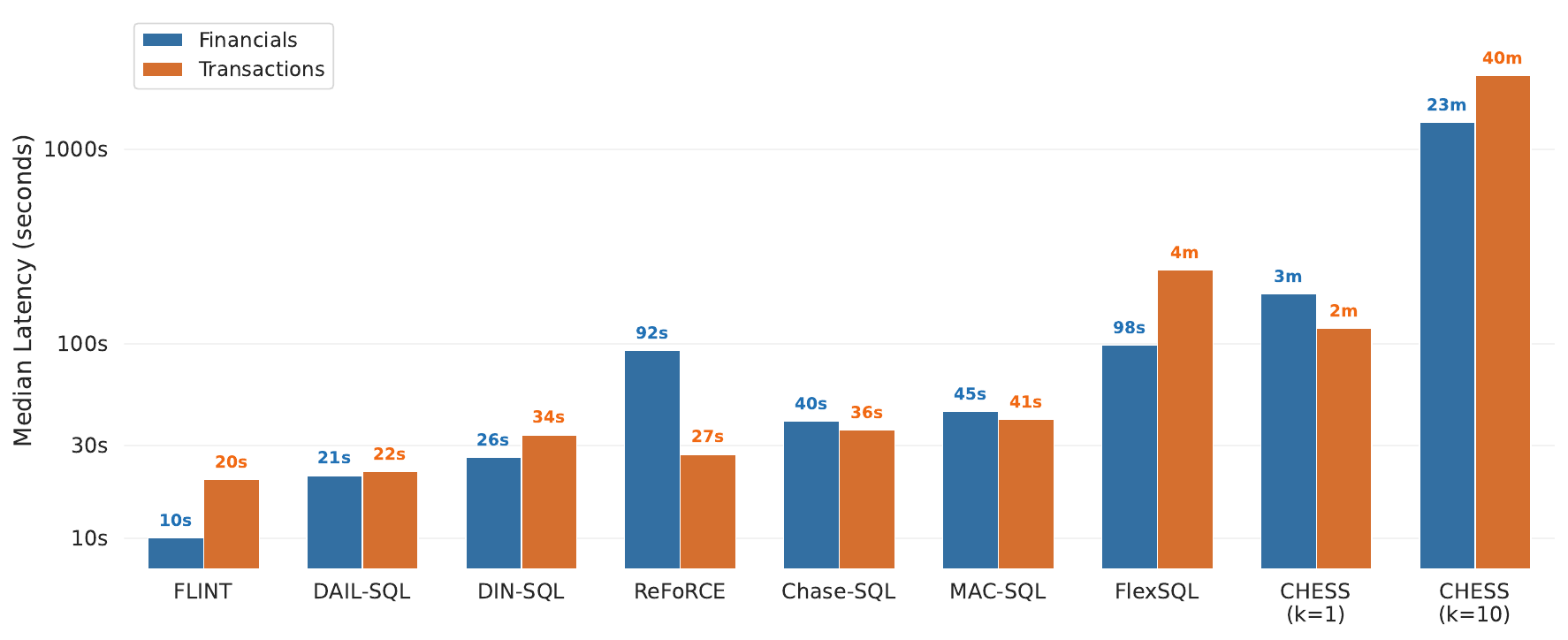}
\caption{\small{Median end-to-end query latency (log scale). \ac{FLINT} achieves 10--20s median, almost 2$\times$ faster than the next-fastest baseline on Financials and almost 100$\times$ faster than CHESS.}}
\label{fig:latency}
\end{figure*}

As a deployed production system, \ac{FLINT}'s end-to-end latency is a critical design constraint.
Figure~\ref{fig:latency} compares median query latency across all methods on a logarithmic scale.

\ac{FLINT}'s median of 10--20s enables interactive production deployment beating all the compared \acs{T2S} systems. CHESS's latency is driven by sequential candidate generation and database exploration, making it infeasible for real-time use.

\section{Ablation Details}
\label{sec:appendix_ablations}

Each ablation disables exactly one component while keeping the rest of the pipeline intact.
This section specifies the experimental condition for each.

\paragraph{Lookup Agent.}
The dynamic Lookup Agent is replaced with a static mapping of the 100 most common Financials IDs and 50 most common Transactions IDs, covering about 78\% of queries.. These mappings are injected into the generation prompt as pre-resolved constraints. This measures whether \emph{question-specific} constraint resolution adds value over comprehensive static knowledge. 

\paragraph{Schema Linking.}
The schema linking stage is bypassed entirely.
The SQL generator receives the full schema (all tables and columns) without intermediate table or column selection.
All other components (lookup, retrieval, entity linking, reflection) remain active.

\paragraph{Schema linking ablation condition.}
When schema linking is removed, the model must select among hundreds of columns from the full schema text.
Table overlap remains high (0.96) because the lookup constraints and retrieved examples still guide table selection implicitly, but column selection degrades.

\paragraph{Few-Shot Retrieval.}
The similar query retrieval node returns zero examples.
Schema linking and SQL generation operate in zero-shot mode---no exemplar queries are provided.
The lookup agent still provides reference-table constraints, isolating the contribution of structural templates.

\paragraph{Reflection.}
The first generated SQL is submitted without retry regardless of execution outcome. On Financials, fewer than 1\% of queries trigger reflection, so this ablation has minimal effect.
On Transactions, 16\% of queries initially fail execution, providing the opportunity for reflection to recover.

\section{Query Bank Overlap Analysis}
\label{sec:appendix_overlap}

We test whether evaluation accuracy could be explained by similarity between the retrieval query bank and the evaluation set. For each evaluation query we compute the normalized tree-edit distance (TED) to its nearest query bank neighbor over canonicalized \texttt{sqlglot} ASTs, in a content-aware mode where table/column names and literals contribute to the distance (the leakage-relevant measure).
Overlap is reported as $1-\text{TED}$.

\begin{table}[h]
\centering
\small
\caption{Nearest-neighbor content-aware overlap ($1-\text{TED}$) from each evaluation query to the query bank.}
\setlength{\tabcolsep}{5pt}
\renewcommand{\arraystretch}{1.2}
\begin{tabular}{@{}l ccc@{}}
\toprule
\textbf{Dataset} & \textbf{Median} & \textbf{$\geq$90\% match} & \textbf{Closest (min TED)} \\
\midrule
Financials  & 0.78 & 10.6\% & 0.023 \\
Transactions & 0.62 & \phantom{0}0.0\% & 0.111 \\
\bottomrule
\end{tabular}
\label{tab:overlap}
\end{table}

Three points show overlap does not drive accuracy. (1) \textbf{Control:} baselines
given the same bank (few-shot $k{=}3$, and DAIL/DIN/FlexSQL/ReFoRCE via the same
retrieval) still trail by ${\sim}20$--$30$ points; if overlap conferred accuracy
they would inherit it. (2) \textbf{Disjoint by construction:} bank examples use a
fixed canonical entity and share no question with the evaluation set, so even a
structurally similar example filters on different entities/metrics/dates and is
not answerable by copying. (3) \textbf{Structural, not answer-level:} the high structural similarity reflects a shared join backbone (a common multi-table chain plus currency handling) by design; the content-aware measure, which is the leak-relevant one, shows essentially no near-duplicates on Transactions.

\section{Generalization to Other Domains}
\label{sec:generalization}
To test whether \ac{FLINT}'s architecture transfers beyond finance, we evaluate on two external benchmarks spanning various scientific and industrial domains. For each, we adapt \ac{FLINT} with only configuration changes, i.e. schema documentation, a query bank for retrieval, and domain-specific prompt rules, without any modification of the pipeline code. Our code is available at \url{https://github.com/kensho-technologies/flint}.

SM3-Text-to-Query~\citep{sivasubramaniam2024smtexttoquery} is a medical benchmark over Synthea synthetic patient data (19 PostgreSQL tables, SNOMED/LOINC/RxNorm codes as opaque identifiers). The high few-shot baseline reflects structural simplicity: 68\% of questions are single-table. \ac{FLINT}'s gains concentrate on the harder slices, i.e. multi-table-join accuracy reaches 97\% and unanswerable-question detection 97\%, versus 90\% and 42\% for few-shot.

ScienceBenchmark~\citep{sciencebenchmark} provides a harder test: three PostgreSQL databases from distinct scientific domains where even few-shot achieves only 50--82\%. CORDIS (EU research funding, 19 tables) requires navigating a complex many-to-many schema linking projects, institutions, and funding programmes. OncoMX (cancer biomarkers, 25 tables) demands correct join paths through a deep foreign-key graph connecting genes, diseases, and clinical tests. SDSS (astronomical surveys, 6 tables, 4.2M rows) tests retrieval over a wide-column photometric catalog with domain-specific type lookups.

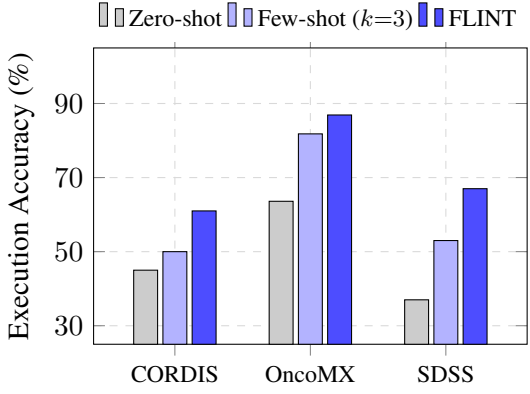
\begin{figure}[t]
\centering
\begin{tikzpicture}
\begin{axis}[
    ybar,
    bar width=9pt,
    width=0.95\columnwidth,
    height=5.5cm,
    ylabel={Execution Accuracy (\%)},
    symbolic x coords={CORDIS, OncoMX, SDSS},
    xtick=data,
    xticklabel style={font=\small},
    ymin=25, ymax=105,
    ytick={30,50,70,90},
    legend style={at={(0.5,1.02)}, anchor=south, legend columns=3, font=\small, draw=none},
    enlarge x limits=0.3,
    grid=major,
    grid style={dashed, gray!30},
]
\addplot[fill=gray!40] coordinates {(CORDIS,45.0) (OncoMX,63.6) (SDSS,37.0)};
\addplot[fill=blue!30] coordinates {(CORDIS,50.0) (OncoMX,81.8) (SDSS,53.0)};
\addplot[fill=blue!70] coordinates {(CORDIS,61.0) (OncoMX,86.9) (SDSS,67.0)};
\legend{Zero-shot, Few-shot ($k{=}3$), \ac{FLINT}}
\end{axis}
\end{tikzpicture}
\caption{
    \small{Accuracy on ScienceBenchmark~\cite{sciencebenchmark}: CORDIS (EU funding, 19 tables), OncoMX (cancer biomarkers, 25 tables), SDSS (astronomy, 6 tables).}
}
\label{fig:sciencebenchmark_results}
\end{figure}

\begin{table}[t]
\centering
\caption{\small{SM3-Text-to-Query~\cite{sivasubramaniam2024smtexttoquery} results. Hard subset: multi-table joins and unanswerable questions. \ac{FLINT}'s gain concentrates on the hard subset (+11pp over few-shot).}}
\small
\setlength{\tabcolsep}{6pt}
\renewcommand{\arraystretch}{1.2}
\begin{tabular}{@{}l rr rr@{}}
\toprule
& \multicolumn{2}{c}{\textbf{SM3-Text-to-Query}} \\
\cmidrule(lr){2-3}
\textbf{Method} & \textit{All (n=2000)} & \textit{Hard (n=631)} \\
\midrule
Zero-shot & 77.8 & 58.2 \\
Few-shot ($k{=}3$) & 92.3 & 85.3 \\
\rowcolor{blue!6}
\textbf{\ac{FLINT}} & \textbf{95.7} & \textbf{96.5} \\
\bottomrule
\end{tabular}
\label{tab:sm3_results}
\end{table}
FLINT improves over few-shot in every domain, by as much as 14 points.(Fig.~\ref{fig:sciencebenchmark_results} and Tab.~\ref{tab:sm3_results}). The gains come from the same components: Retrieval demonstrates correct join patterns for the domain's schema topology, domain rules enforce conventions (case sensitivity, operator choice, table-specific semantics), and execution-based reflection recovers from wrong join paths by detecting empty results.
No architectural changes are required; only the configuration layer adapts to each domain.

\section{Deployment Considerations}
\label{sec:appendix_deployment}

\ac{FLINT} serves production traffic against a multi-terabyte financial database. Unlike offline evaluation, deployment has no gold queries, so the system must ensure safety and detect failures without ground truth supervision.

\subsection{Query Safety}

All generated SQL executes through a read-only connection with row-level access controls. Queries are rejected at the gateway if they reference tables or columns outside the user's entitlement scope.
A structural validator inspects every query before execution, blocking statements that contain mutation keywords, unbounded scans without \texttt{WHERE} clauses, or joins exceeding a configurable fan-out threshold. Query execution is subject to per-session resource limits (timeout, memory, result-set size) enforced by the warehouse, preventing runaway queries from degrading shared infrastructure.

\subsection{Monitoring and Quality Signals}

Without ground truth, we rely on proxy signals to detect degradation.
Execution failures (syntax errors, permission violations, timeouts) provide an unambiguous error signal and trigger immediate retry via the reflection loop. For semantically wrong but executable queries, we monitor result set anomalies: empty results for questions that should return data, unexpectedly large result sets, and null ratios inconsistent with the queried tables. User feedback (explicit ratings and implicit signals such as re-asking a rephrased question) provides a sparse but high-quality correctness signal. Aggregate metrics (execution success rate, median latency, and user satisfaction scores) are tracked over rolling windows; sustained degradation triggers review of the retrieval index and prompt templates.

\section{Expanded Related Work}
\label{sec:appendix:expanded_related_work}

Outside of multi-agent systems, earlier prompt-engineering approaches tackle the problem without multi-agent orchestration.
DAIL-SQL~\cite{gao2024dail} uses token-level similarity for few-shot selection with a structured format separating schema from examples, while DIN-SQL~\cite{pourreza2024din} decomposes the task into sub-problems with self-correction. 
Both methods indicate that prompt structure matters, but neither includes mechanisms for finding values that appear in the question or schema text (the Lookup Agent~\ref{sec:lookup} in \ac{FLINT}).

A separate line of work focuses on schema-linking as a pre-processing step to reduce the amount of tokens presented to the SQL generator.
C3-SQL~\cite{dong2023c3zeroshottexttosqlchatgpt} uses ChatGPT to identify relevant schemas through zero-reasoning.
RESDSQL~\cite{resdsql} trains a separate ranking model to select relevant tables and columns to then pass a pruned schema to the generator. 
On normalized schemas, we have the extra challenge of not only identifying target tables (where the answer resides), but also bridge tables that are used for foreign-key traversal.
\ac{FLINT} addresses this by using join paths, resolved lookup constraints, and similar retrieved examples to guide bridge-table identification.

\end{document}